\pdfoutput=1

\documentclass[11pt]{article}

\usepackage[final]{coling}

\usepackage{times}
\usepackage{latexsym}

\usepackage[T1]{fontenc}

\usepackage[utf8]{inputenc}

\usepackage{microtype}

\usepackage{inconsolata}
\usepackage{makecell}
\usepackage{multirow}
\usepackage{colortbl}
\usepackage{amsmath}
\usepackage{amssymb}
\usepackage{booktabs}
\usepackage{graphicx}

\title{Multi-View Incongruity Learning for Multimodal Sarcasm Detection}

\author{
 \textbf{Diandian Guo\textsuperscript{1,2}},
 \textbf{Cong Cao\textsuperscript{1,}\thanks{Corresponding authors}},
 \textbf{Fangfang Yuan\textsuperscript{1}},
 \textbf{Yanbing Liu\textsuperscript{1,2,*}},
\\
 \textbf{Guangjie Zeng\textsuperscript{3}},
 \textbf{Xiaoyan Yu\textsuperscript{4}},
 \textbf{Hao Peng\textsuperscript{3}},
 \textbf{Philip S. Yu\textsuperscript{5}}
\\
 \textsuperscript{1}Institute of Information Engineering, Chinese Academy of Sciences\\
 \textsuperscript{2}School of Cyber Security, University of Chinese Academy of Sciences\\
 \textsuperscript{3}State Key Laboratory of Software Development Environment, Beihang University\\
 \textsuperscript{4}School of Computer Science and Technology, Beijing Institute of Technology\\
 \textsuperscript{5}Department of Computer Science, University of Illinois at Chicago
\\
 {\texttt{\{guodiandian, caocong, yuanfangfang, liuyanbing\}@iie.ac.cn},} \\ { \texttt{\{zengguangjie,penghao\}@buaa.edu.cn,}\texttt{xiaoyan.yu@bit.edu.cn, psyu@uic.edu}}
 }

\begin{document}
\maketitle

\begin{abstract}
Multimodal sarcasm detection (MSD) is essential for various downstream tasks.
Existing MSD methods tend to rely on spurious correlations. 
These methods often mistakenly prioritize non-essential features yet still make correct predictions, demonstrating poor generalizability beyond training environments. 
Regarding this phenomenon, this paper undertakes several initiatives.
Firstly, we identify two primary causes that lead to the reliance of spurious correlations. 
Secondly, we address these challenges by proposing a novel method that integrate \textbf{M}ultimodal \textbf{I}ncongruities via \textbf{C}ontrastive \textbf{L}earning (MICL) for multimodal sarcasm detection. 
Specifically, we first leverage incongruity to drive multi-view learning from three views: token-patch, entity-object, and sentiment. 
Then, we introduce extensive data augmentation to mitigate the biased learning of the textual modality. 
Additionally, we construct a test set, SPMSD, which consists potential spurious correlations to evaluate the the model's generalizability. 
Experimental results demonstrate the superiority of MICL on benchmark datasets, along with the analyses showcasing MICL's advancement in mitigating the effect of spurious correlation.
\end{abstract}

\section{Introduction}

Sarcasm, inherently metaphorical, seeks to convey meanings that diverge from literal interpretations. 
Its prevalence on social media platforms underscores the critical need for effective sarcasm detection, which is a tool pivotal for uncovering the genuine opinions and emotions of users. 
This capability supports essential applications such as public opinion mining \cite{cai2019multi, prasanna2023polarity} %
and sentiment analysis \cite{farias2017irony, khare2023sentiment}. %

\begin{figure}[t]
  \centering
  \includegraphics[width=\linewidth]{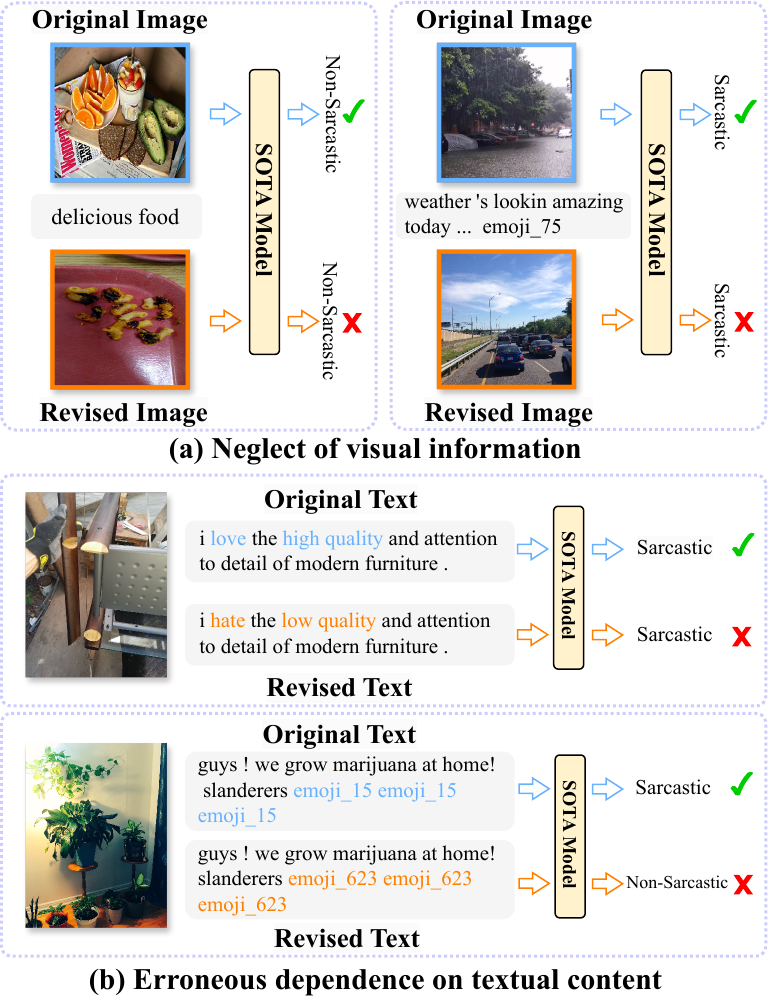}
  \caption{Existing models suffer from two deficiencies that lead to spurious correlations on MSD task.}
  \label{fig:mot}
\end{figure} 

Early attempts of sarcasm detection focus solely on textual modality \cite{davidov2010semi, zhang2016tweet, xiong2019sarcasm} %
, modeling the incongruities within the text. %
However, the proliferation of multimedia social platforms enables users to convey opinions and emotions using multimodal information.
As a result, MSD has recently attracted widespread attention.
\citet{joshi2015harnessing} demonstrate the incongruity as a pivotal factor for detecting sarcasm, which sparks a surge of research into learning incongruity using textual and visual cues, achieving outstanding results \cite{ wen2023dip, qiao2023mutual}.

Despite these efforts, existing models still suffer from reliance on spurious correlations. 
Spurious correlation is a phenomenon where models learn non-generalizable features, rather than core features truly related to the real labels, thus undermining the model's generalizability \cite{deng2024robust}. 
We conduct experiments on the current SOTA model \cite{jia2023debiasing} and attribute the spurious correlations in MSD to two primary oversights: %
\textbf{1)} Overemphasis on the text encoder while underestimating visual information. For example shown in Figure \ref{fig:mot}(a), changing the image does not affect the model's result when the textual input remains the same, revealing a biased dependence on the textual modality.
\textbf{2)} Erroneously relying on non-critical textual features rather than critical emotional features.  As the instance illustrated in Figure \ref{fig:mot}(b), changing key emotional words does not affect the model's result. Conversely, the model makes an opposite judgment when non-critical descriptions that do not affect semantics are modified. In summary, the above findings reveal that existing models rely on spurious correlations, failing to capture the necessary task-related features.

To address the above issues, we introduce MICL, a novel multi-view incongruity learning method for MSD. This method is structured around three modules:  multimodal feature encoding, multi-view incongruity learning, and multi-view fusion. 
Specifically, for multimodal feature encoding, in addition to the traditional textual and visual encoding, we introduce the OCR-texts for supplementary element to uncover the information contained within the image to a greater extend. 
\citet{yang2024sebot} demonstrate that multi-view learning can improve the effectiveness of models in social media.
Considering that sarcastic content often involves an entity or object in a multimodal context and carries sentiment polarity, the multi-view incongruity learning module learns robust features from three aspects: token-patch, entity-object, and sentiment, to mitigate spurious correlations. 
However, the quality and importance of each view vary significantly across samples \cite{wu2022characterizing}.
Therefore, we propose using a beta distribution-based multi-view fusion module to perform confidence-weighted fusion of the learned embeddings, producing more reliable results. %
Furthermore, we extend beyond conventional text data augmentation techniques, which tend to perpetuate a bias towards textual information. Instead, MICL incorporates a dual augmentation strategy, enhancing both text and image data. %
Our contributions are as follows.

$\bullet$ \; We propose MICL, a novel multi-view learning method that comprehensively learns incongruities and integrates them credibly. %

$\bullet$ \; %
We introduce robust data augmentation strategies that enriches both textual and visual contents, mitigating biased learning of the textual modality.

$\bullet$ \; Experimental results indicate that our approach outperforms existing methods on the MSD task and demonstrates stronger robustness against spurious correlations.

\begin{figure*}[h]
  \centering
  \includegraphics[width=\linewidth]{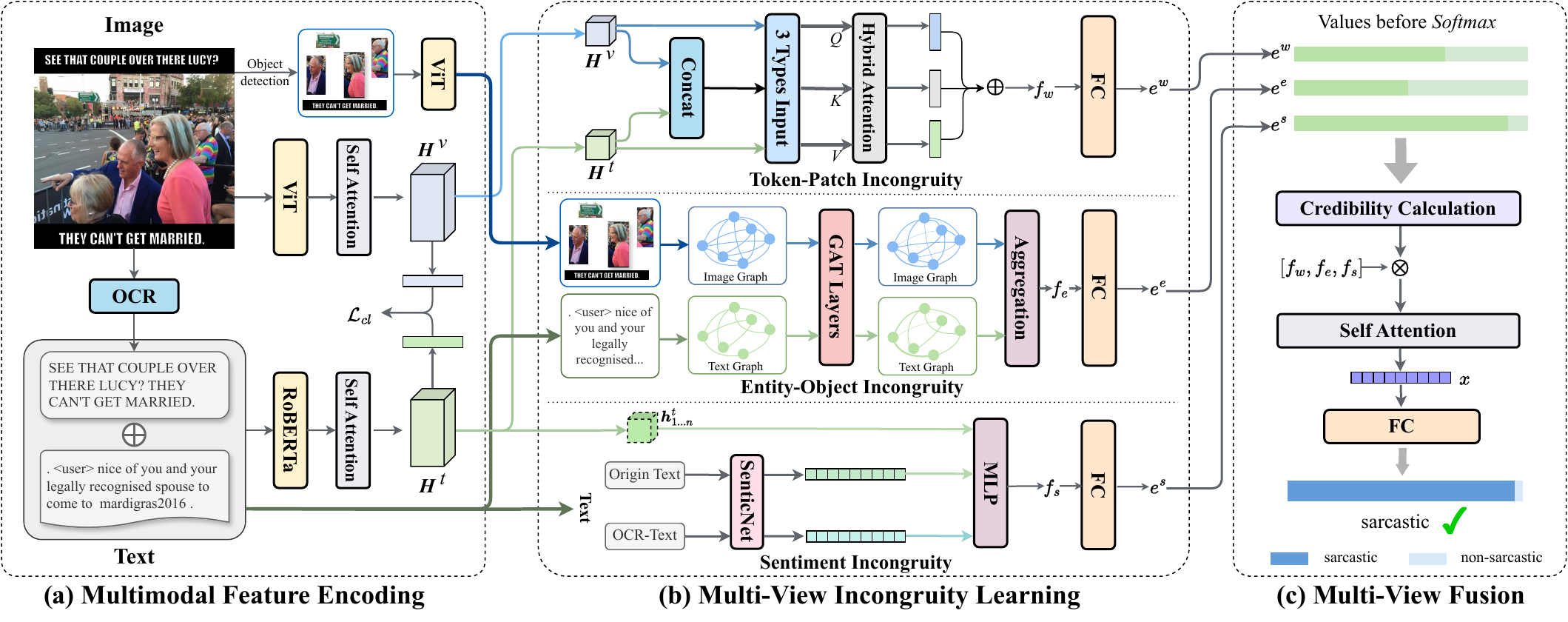}
  \caption{The overall architecture of MICL primarily comprises three key modules: (a) Multimodal Feature Encoding, (b) Multi-View Incongruity Learning, and (c) Multi-View Fusion. Additionally, we introduce data augmentation for each training data.}
  \label{fig:model}
\end{figure*}

\section{Related Work}
\subsection{Multimodal Sarcasm Detection}
Multimodal sarcasm detection is a research task that  identifies sarcasm through multimodal cues. 
\citet{schifanella2016detecting} first propose integrating textual features with visual features to solve the sarcasm detection task. Following this,  \citet{cai2019multi} construct an advanced MSD dataset based on tweets, providing a benchmark for subsequent research. %
InCrossMGs  \cite{liang2021multi} is the first to model the interaction of information within and between modalities by graph neural networks. %
DMSD-CL  \cite{jia2023debiasing} employs counterfactual augmentation and contrastive learning to study MSD in out-of-distribution scenarios. Recently, many works have dedicated efforts to model the incongruity in text-image pairs. For example, %
MILNet \cite{qiao2023mutual} focuses on the combination of local incongruity and global incongruity. %
However, existing methods only focus on token-patch incongruity, which leads to erroneous reliance on non-critical features. Our model proposes to learn multi-view incongruity information to improve performance and enhance robustness. %

\subsection{Mitigating Spurious Correlations}
Mitigating spurious correlations in multimodal scenarios has attracted increasing research interest. Existing methods for improving robustness against spurious correlations can be divided into two lines of research. One line focuses on effectively using multimodal information to enhance robustness \cite{Yenamandra_2023_ICCV}. Some methods use the distributed robust optimization (DRO) framework to dynamically increase the weight of minimizing the worst group loss \cite{wen2021toward}. Most recently, \citet{kirichenko2023last} propose methods that train a model using Empirical Risk Minimization (ERM) first and then only finetune the last layer on balanced data. Another line of research focuses on mitigating the bias in training data by creating additional data to balance the training dataset \cite{niu2021counterfactual}. Inspired by these methods, we comprehensively mitigate the reliance on spurious correlations in the MSD task from both the model and data perspectives.

\section{Methodology}
As shown in Figure \ref{fig:model}, the architecture of MICL mainly consists of three parts: multimodal feature encoding, multi-view incongruity learning, and multi-view fusion. Additionally, to mitigate the modal bias problem from the data level, we introduce data augmentation for the training input.

\subsection{Multimodal Feature Encoding}
Given a text-image pair $\mathcal{X}=(\mathcal{T},\mathcal{V})$, we first need to perform feature encoding, which is divided into two steps: text encoding and image encoding.
\subsubsection{Text Encoding}
 
\begin{figure}[htbp]
  \centering
  \includegraphics[width=\linewidth]{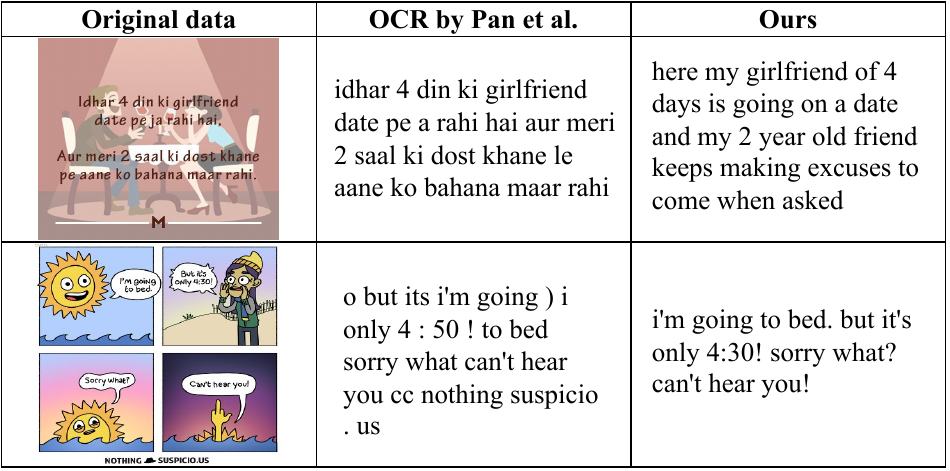}
  \caption{In the first example, since the text is in Hindi, it is difficult for a non-multilingual pre-trained RoBERTa to understand. Our method automatically translates the extracted text into English. In the second example, existing OCR result exhibits deficiencies in both recognition accuracy and sequential integrity, whereas our result performs better.}
  \label{fig:ocr}
\end{figure}

In current multimodal learning approaches, textual and visual information are commonly encoded independently. However, our observation reveals that a number of images contain textual information that frequently complements the textual modality. 
Building upon this observation, we incorporate optical character recognition text (OCR-text) $\mathcal{O}$ from images as an auxiliary input alongside the original text input $\mathcal{T}$. 
However, the OCR-text provided by existing work \cite{pan2020modeling} has issues with low accuracy and ambiguous meaning, as shown in the Figure \ref{fig:ocr}. Low-quality OCR-text may reduce model performance \cite{wang2024generated}. 
Instead, we generate refined OCR-texts employing GLM-4V\footnote{\url{https://open.bigmodel.cn}}  \cite{wang2023cogvlm} with more precision extraction and translation, complemented by meticulous manual proofreading.
Then, we concatenate $\mathcal{T}$ and $\mathcal{O}$, and feed them into the text encoder. As shown in Figure \ref{fig:model}(a), we apply the pre-trained language model RoBERTa \cite{liu2019roberta} as the text encoder:
\begin{equation}
    \boldsymbol{H}^t =  \mathrm{Self\_Att}( \mathrm{RoBERTa}(\mathcal{T}\oplus \mathcal{O})) , %
\end{equation}
where $\boldsymbol{H}^t = [\boldsymbol{h}_{cls}^t,\boldsymbol{h}_1^t,\boldsymbol{h}_2^t,...,\boldsymbol{h}_{n}^t] \in \mathbb{R}^{(n+1) \times d }$ is the textual representation of the input text, $\boldsymbol{h}^t_i \in \mathbb{R}^d $ denotes the hidden state vector of $i$-token, $d$ denotes the dimension of the hidden representations, $n$ is the total number of tokens after concatenating the original text and OCR-text, $\mathrm{Self\_Att}$ means a self-attention layer, and $\oplus$ refers to the concatenation operation. For clarity and simplification, we use $\boldsymbol{e}_t^k$ to represent $\boldsymbol{h}_{cls}^t$ of the $k$-th sample in subsequent expressions.

\subsubsection{Image Encoding}
We use a pre-trained ViT \cite{dosovitskiy2020image} as the image encoder. For each image $\mathcal{V}=\{v_{cls},v_1,...,v_{n_\mathcal{V}}\}$, where $v_{cls}$ means the \verb|[CLS]| token, $v_i$ represents the $i$-patch of $\mathcal{V}$, and $n_\mathcal{V}$ is the total patch number. We feed $\mathcal{V}$ into ViT:
\begin{equation}
    \boldsymbol{H}^v =\mathrm{Self\_Att}(\mathrm{ViT}(\mathcal{V})) ,
\end{equation}
where $\boldsymbol{H}^v = [\boldsymbol{h}_{cls}^v,\boldsymbol{h}_1^v,\boldsymbol{h}_2^v,...,\boldsymbol{h}_{n_\mathcal{V}}^v] \in \mathbb{R}^{(n_\mathcal{V}+1) \times d }$ is the visual representation of the input image, $\boldsymbol{h}_i^v \in \mathbb{R}^d$ represents the $i$-th patch embedding. For clarity and simplification, we use $\boldsymbol{e}_v^k$ to represent $\boldsymbol{h}_{cls}^v$ of the $k$-th sample in subsequent expressions.

\subsection{Multi-View Incongruity Learning}

For the MSD task, cross-modal incongruity learning predominantly focuses on the token-patch levels. 
However, sarcastic contents are often closely related to specific entities or objects in multimodal contexts. 
Furthermore, sarcastic contents typically involve strong emotions that existing models overlook. 
To achieve a more comprehensive incongruity learning, we further incorporate incongruity learning from the entity-object and the explicit sentiment perspectives as shown in Figure \ref{fig:model}(b).

\subsubsection{Token-patch Incongruity Learning}
A cross-attention mechanism is commonly used for modeling cross-modal interactions. Existing methods \cite{qiao2023mutual,jia2023debiasing} often use text as the query and images as the key and value, which may lead to modality bias. %
Instead, we design a hybrid attention interaction mechanism for unbiased token-patch incongruity learning, which can integrate text and image features in a balanced manner. 
Based on the input differences of the multi-head attention layer, it can be divided into the following parts:
\begin{gather}
    \boldsymbol{Q}_{tv}=\boldsymbol{K}_{tv}=\boldsymbol{V}_{tv}=\boldsymbol{H}^{tv} \; , \label{eq:1}  \\
    \boldsymbol{Q}_t=\boldsymbol{H}^t, \boldsymbol{K}_t=\boldsymbol{V}_t=\boldsymbol{H}^v \; ,  \label{eq:2} \\
    \boldsymbol{Q}_v=\boldsymbol{H}^v, \boldsymbol{K}_v=\boldsymbol{V}_v=\boldsymbol{H}^t \; , \label{eq:3}
\end{gather}
where $\boldsymbol{H}^{tv}=\boldsymbol{H}^t\oplus \boldsymbol{H}^v$. Then, we feed different inputs into a standard cross-attention layer:
\begin{gather}
     \boldsymbol{F} =  \mathrm{Cross\_att}(\boldsymbol{Q},\boldsymbol{K},\boldsymbol{V}).
\end{gather}

We define $\boldsymbol{F}_{tv}$, $\boldsymbol{F}_{t}$ and $\boldsymbol{F}_{v}$ as the outputs of the attention mechanisms from the input Eq. \eqref{eq:1}, \eqref{eq:2} and \eqref{eq:3}, respectively. For $\boldsymbol{F}_{tv}$, $\boldsymbol{F}_{t}$ and $\boldsymbol{F}_{v}$, we treat the encoding of their \verb|[CLS]| tokens, $\boldsymbol{f}_{tv}$, $\boldsymbol{f}_{t}$ and $\boldsymbol{f}_{v}$, as the final output:
\begin{equation}
    \boldsymbol{f}_{w} = \boldsymbol{f}_{tv} \oplus \boldsymbol{f}_{t} \oplus \boldsymbol{f}_{v}\;.
\end{equation}

\subsubsection{Entity-object Incongruity Learning}
To effectively capture entity-object incongruity, we construct semantic graphs for both text and images. Specifically, for the text semantic graph, we treat entities as nodes and use spaCy\footnote{\url{https://spacy.io/}} to extract dependencies between entities as edges. If there is a dependency between two entites, an edge will be created between them in the text graph. For the visual semantic graph, we follow \citet{anderson2018bottom} to segment the image into object regions. We treat each region as a node, and create edges based on cosine similarity. Additionally, both graphs are undirected and contain self-loops.

Then, we model the graphs with Graph Attention Network (GAT) \cite{velivckovic2017graph}. %
Taking the textual graph as an example, let $\alpha^l_{i,j}$ be the attention score between $i$ and $j$, and $\boldsymbol{g}_i^l$ denote the feature of node $i$ in the $l$-th layer. We have:
\begin{gather}
\alpha_{i,j}^l=\frac{\exp\left(LR\left(\boldsymbol{u}_l^\top[\boldsymbol{W}_l\boldsymbol{g}_i^l\|\boldsymbol{W}_l\boldsymbol{g}_j^l]\right)\right)}{\sum_k\exp\left(LR\left(\boldsymbol{u}_l^\top[\boldsymbol{W}_l\boldsymbol{g}_i^l\|\boldsymbol{W}_l\boldsymbol{g}_k^l]\right)\right)}, \\
\boldsymbol{g}_i^{l+1}=\alpha_{i,i}^l\boldsymbol{W}_l\boldsymbol{g}_i^l+\sum_{j\in\mathcal{N}(i)}\alpha_{i,j}\boldsymbol{W}_l\boldsymbol{g}_j^l \;\;,
\end{gather}
where $k \in \mathcal{N}(i) \cup i$ belongs to the neighbor nodes of $i$ and $i$ itself. $LR$ denotes the $\mathrm{LeakyReLU}$ layer. $\boldsymbol{W}_l \in \mathbb{R}^{d\times d}$ and $\boldsymbol{u}_l$ are learnable parameters of the $l$-th textual GAT layer. %
We initialize $\boldsymbol{g}_i^0 = \boldsymbol{h}_i^t$.

We denote the final textual representation as $\boldsymbol{G}^T = \{\boldsymbol{g}_0,...,\boldsymbol{g}_n \}$. Similarly, we can obtain $\boldsymbol{G}^V$. We define $\boldsymbol{G}=\boldsymbol{G}^T \oplus \boldsymbol{G}^V$, then we can learn the entity-object incongruity:
\begin{gather}
        \boldsymbol{f}_{e} = \frac{1}{|\boldsymbol{G}|}\sum_{\boldsymbol{g}_i\in \boldsymbol{G}} \mathrm{Softmax}\left(\boldsymbol{g}_i \boldsymbol{W}_g + b_g\right) \boldsymbol{g}_i ,
\end{gather}
where $\boldsymbol{W}_g $ and $b_g$ are learnable parameters.

\subsubsection{Sentiment Incongruity Learning}

Given the pivotal role of emotional context in MSD \cite{joshi2015harnessing}, our model integrates sentiment analysis to discern incongruity in the original text and OCR-text. Specifically, we extract the sentiment polarity of the source text and OCR-text through SenticNet \cite{cambria2024senticnet}:
\begin{gather}
    \boldsymbol{s}_t=\mathrm{SenticNet}(\mathcal{T}), \; \boldsymbol{s}_o=\mathrm{SenticNet}(\mathcal{O}) , \\
    \boldsymbol{f}_{s} = \mathrm{MLP}(\boldsymbol{s}_t  \oplus \boldsymbol{s}_o \oplus \boldsymbol{h}_{1...n}^t) ,
\end{gather}
where $\mathrm{MLP}$ is a muti-layer perceptron. If OCR-text is unavailable, $\boldsymbol{f}_{s}$ is assigned a value of 0.

\begin{figure}[htbp]
  \centering
  \includegraphics[width=\linewidth]{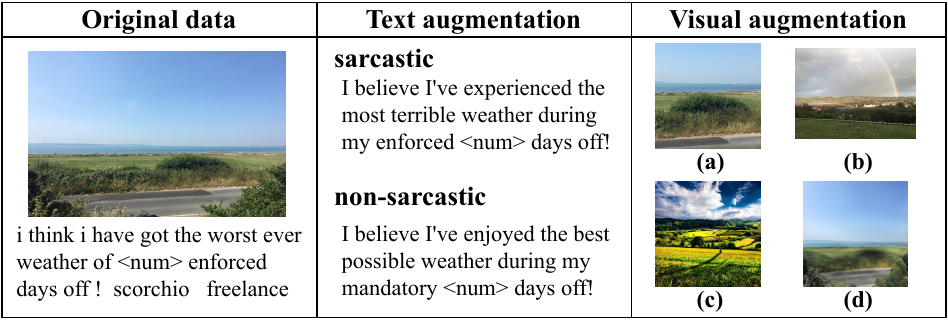}
  \caption{Summary of text and visual augmentation methods. Text augmentation generates samples with the same or opposite labels. Visual augmentation methods include: (a) cropping, (b) swapping images, (c) image generation, and (d) image style transfer.}
  \label{fig:aug}
\end{figure}

\subsection{Multi-View Fusion}
As shown in Figure \ref{fig:model}(c), the credibility of the three incongruity features varies across different MSD scenarios. Measuring the confidence of different features helps improve detection performance. TMC \cite{han2020trusted} has proved that the Dirichlet distribution can effectively estimate the credibility of a single view. In binary classification scenario, the Beta distribution shares the same mathematical significance.  Following \citet{ma2024event}, we use the output before the softmax operation of the $m$-th view classifier as evidence $e^m$, then the credibility $c^m$ can be expressed as: 
\begin{equation}
    c^m = \frac{e^m_0+e^m_1}{S_m} = \frac{e^m_0+e^m_1}{(e^m_0+1)+(e^m_1+1)} ,
\end{equation}
where $e^m_r$ represents the output of the final layer of the classifier for the $m$-th view regarding the $r$-th classification result. In binary classification tasks, $r\in\{0,1\}$. The derivation process can be found in the Appendix \ref{sec:cre}.

After obtaining the credibility, we use a self-attention network to obtain the fusion feature:
\begin{equation}
    \boldsymbol{x} = \mathrm{Self\_Att}([\boldsymbol{f}_{w},\boldsymbol{f}_{e},\boldsymbol{f}_{s}]\cdot[c^w,c^e,c^s]^\top).
\end{equation}

\subsection{Data Augmentation and Contrastive Learning}

\subsubsection{Data Augmentation}
Images serve as a vital source of incongruity clues, which is essential for comprehensive sarcasm analysis. 
However, previous MSD methods \cite{pan2020modeling,jia2023debiasing} focus on enhancing textual content and overlook the importance of image data augmentation. 
This inadequate data augmentation fails to enhance model performance and may even impede the performance \cite{wang2024generated}. 
To address this issue, we adopt augmentation involving both textual and visual data, ensuring a balanced and effective enhancement. 

As shown in Figure \ref{fig:aug}, for text augmentation, we employ two strategies: \textbf{1)} Replacing key entities or reversing sentiment words to obtain samples with opposite labels; \textbf{2)} Paraphrasing the original samples to keep the meaning unchanged, obtaining samples with the same labels. Text augmentation is performed by ChatGPT\footnote{\url{https://chat.openai.com}}. We apply the above strategies at a 1:1 ratio to generate augmented texts for all training samples.

For image augmentation, we use four strategies: \textbf{1)} Randomly cropping images and resizing them to 224$\times$224; \textbf{2)} Randomly swapping images of samples with the same label; \textbf{3)} Employing stable diffusion for image style transfer; \textbf{4)} Prompting GLM-4V to generate image titles, and then using stable diffusion to generate new images based on those titles.  We apply these four strategies at a 3:3:2:2 ratio to generate augmented images for all training samples.

\begin{table*}[h]
\centering
\
\caption{Main results on MMSD dataset for sarcasm detection. We use $^*$ indicates the reproduced results by using RoBERTa as the textual backbone.}
\label{tab:main}
\resizebox{\linewidth}{!}{  
\begin{tabular}{clccccccc}
 \Xhline{1.2pt}
\multirow{2}{*}{Modality}    & \multicolumn{1}{l}{\multirow{2}{*}{Method}} & \multirow{2}{*}{Acc.(\%)} & \multicolumn{3}{c}{Binary-Average}                                                     & \multicolumn{3}{c}{Macro-Average}                                                      \\ \cline{4-9} 
                             & \multicolumn{1}{c}{}                        &                          & \multicolumn{1}{l}{Pre.(\%)} & \multicolumn{1}{l}{Rec.(\%)} & \multicolumn{1}{l}{F1(\%)} & \multicolumn{1}{l}{Pre.(\%)} & \multicolumn{1}{l}{Rec.(\%)} & \multicolumn{1}{l}{F1(\%)} \\ \hline 
\multirow{4}{*}{Text}        & TextCNN                                     & 80.03                    & 74.29                       & 76.39                       & 75.32                      & 78.03                       & 78.28                       & 78.15                      \\
                             & Bi-LSTM                                     & 81.90                    & 76.66                       & 78.42                       & 77.53                      & 80.97                       & 80.13                       & 80.55                      \\
                             & BERT                                       & 83.85                    & 78.72                       & 82.27                       & 80.22                      & 81.31                       & 80.87                       & 81.09                      \\
                             & RoBERTa                                    & 85.51                    & 78.24                       & 88.11                       & 82.88                      & 84.83                       & 85.95                       & 85.16                      \\ \hline %
\multirow{2}{*}{Image}       & Image                                      & 64.76                    & 54.41                       & 70.80                       & 61.53                      & 60.12                       & 73.08                       & 65.97                      \\
                             & ViT                                        & 67.83                    & 57.93                       & 70.07                       & 63.43                      & 65.68                       & 71.35                       & 68.40                      \\ \hline %
\multirow{11}{*}{Text+Image} & HFM                                        & 83.44                    & 76.57                       & 84.15                       & 80.18                      & 79.40                       & 82.45                       & 80.90                      \\
                            &D\&RNet &84.02 &77.97 &83.42 &80.60 &- &-&-\\

                             & Res-BERT                                   & 84.80                    & 77.80                       & 84.15                       & 80.85                      & 78.87                       & 84.46                       & 81.57                      \\
                             & Att-BERT                                 & 86.05                    & 78.63                       & 83.31                       & 80.90                      & 80.87                       & 85.08                       & 82.92                      \\
                             & CMGCN                                     & 87.55                    & 83.63                       & 84.69                       & 84.16                      & 87.02                       & 86.97                       & 87.00                      \\
                             & Multi-View CLIP                           & 88.33                    & 82.66                       & 88.65                       & 85.55                      & -                           & -                           & - \\
                             & MILNet                                      & 89.50                    & 85.16                       & 89.16                       & 87.11                      & 88.88                       & 89.44                       & 89.12                      
                              \\
                             & DMSD-CL                                     & 88.95                    & 84.89                       & 87.90                       & 86.37                      & 88.35                       & 88.77                       & 88.54                      \\ 
                             & G$^2$SAM$^*$  & 91.07 & 88.27 & 90.09& 89.17 & 90.67 & 90.92 & 90.78 \\
                             
                             &  \cellcolor{gray!20}\textbf{MICL (ours)}                                  & \cellcolor{gray!20}\textbf{92.08}           & \cellcolor{gray!20}\textbf{90.05}              & \cellcolor{gray!20}\textbf{90.61}         & \cellcolor{gray!20}\textbf{90.33}             & \cellcolor{gray!20}\textbf{91.85}              & \cellcolor{gray!20}\textbf{91.77}              & \cellcolor{gray!20}\textbf{91.81}             \\ \Xhline{1.2pt}
\end{tabular}}
\end{table*}
\subsubsection{Contrastive Learning Framework}
For the generated sample $\tilde{\mathcal{X}}=(\tilde{\mathcal{T}},\tilde{V})$, we input $\{\mathcal{X},\tilde{\mathcal{X}}\}$ into the training process together. We construct a contrastive learning framework based on whether the labels are the same to determine the positive and negative examples. Specifically, within a batch, samples with the same label as the anchor sample are considered positive samples, forming the positive sample set $S_{P}$; otherwise, they belong to the negative sample set $S_{N}$. We define the sample set in one batch as $S = S_P+S_N$. In our entire model framework, the key is modeling the incongruity between text and image. Therefore, when constructing the contrastive learning framework, we use the text-image matching approach to obtain scores for positive and negative examples.

For $k$-th sample in the training set, $t \rightarrow v$ contrastive loss is:
\begin{equation}
    \resizebox{\hsize}{!}{$\mathcal{L}^{t \rightarrow v }_{k} = \frac{1}{S_{P}} \sum_{i \in |S_{P}|} - \log \frac{\exp{(\cos(\boldsymbol{e}_t^k,\boldsymbol{e}_v^i)/\tau)}}{\sum_{j \in {S}} \exp{(\cos(\boldsymbol{e}_t^k,\boldsymbol{e}_v^j)/\tau)}} $}, 
\end{equation}
where $\tau \in \mathbb{R}^+$ is the temperature parameter.
Similarly, we can obtain $v \rightarrow t$ contrastive loss $\mathcal{L}^{v \rightarrow t}_{k}$. The overall contrastive loss is as follows:
\begin{equation}
    \mathcal{L}_{cl} = \frac{1}{N} \sum_{k=1}^N {\left(\frac{1}{2}\mathcal{L}^{t \rightarrow v }_{k} +\frac{1}{2}\mathcal{L}^{v \rightarrow t}_{k}\right)}, \label{eq:cl}
\end{equation}
where $N$ is the total number of samples in the training set.

\subsection{Training and Inference}
We obtain the final results based on the fused features:
\begin{equation}
    \hat{y} = \boldsymbol{W}\boldsymbol{x} + b,
\end{equation}
where $\boldsymbol{W}$ and $b$ are learnable parameters. The binary cross-entropy loss is calculated as:
\begin{equation}
    \mathcal{L}_{ce} = -(y\log(\hat{y})+(1-y)\log(1-\hat{y})) .
    \label{eq:ce}
\end{equation}

The final loss function for MICL is defined as the combination of the contrastive learning loss in Eq. \eqref{eq:cl} and the cross-entropy loss in Eq. \eqref{eq:ce}:
\begin{equation}
    \mathcal{L} = \mathcal{L}_{ce} + \lambda \mathcal{L}_{cl} ,
\end{equation}
where $\lambda$ is hyperparameter.

\section{Experiments}
\subsection{Datasets and Metrics}
Our experiments are conducted on the public Multimodal Sarcasm Detection Dataset (MMSD) \cite{cai2019multi}. Each entry in this dataset is a text-image pair, categorized into either sarcastic or non-sarcastic examples based on the specific hashtags. The dataset is divided into a training set, a validating set, and a test set, which includes 19,816, 2,410, and 2,409 samples, respectively. Following previous works \cite{jia2023debiasing}, we report the accuracy, precision, recall, F1-score, and macro-average results to measure the model performance.

To further investigate the models' capability to generalize and their susceptibility to spurious correlations, we meticulously design a small test set, SPMSD. It is refined and expanded from the MMSD dataset, comprising a total of 1,000 samples, including 573 sarcastic items and 427 non-sarcastic items. Detailed information of this dataset can be found in the Appendix \ref{sec:spmsd}.%

\subsection{Baseline Models}
We compare our proposed model MICL with several baselines, which can be broadly categorized into two groups:

\textbf{Unimodal Baselines. }These methods simply take textual or visual information as input, including: TextCNN \cite{kim2014convolutional}, Bi-LSTM \cite{graves2005framewise}, BERT \cite{devlin2018bert} and RoBERTa \cite{liu2019roberta} for textual, Image \cite{cai2019multi} and ViT \cite{dosovitskiy2020image} for visual.

\textbf{Multimodal Baselines. }These methods exploit both visual and textual information as input, including: HFM \cite{ cai2019multi}, D\&RNet \cite{xu2020reasoning}, Res-BERT \cite{pan2020modeling}, Att-BERT \cite{pan2020modeling},  CMGCN \cite{liang2022multi},  Multi-View CLIP \cite{qin2023mmsd2}, MILNet \cite{qiao2023mutual}, DMSD-CL \cite{jia2023debiasing} and G$^2$SAM \cite{wei2024g2sam}.

\begin{table}[htbp]
\small
\centering
\caption{Comparison results on SPMSD dataset (\%).}
\label{tab:ood}
\resizebox{\linewidth}{!}{  
\renewcommand\arraystretch{1}
\begin{tabular}{lcccccccc}
\Xhline{1.2pt}
\multirow{2}{*}{Method}                                                   & \multirow{2}{*}{Acc.}                                                & \multicolumn{3}{c}{Binary-Average}                                                                                                                                                                              & \multicolumn{3}{c}{Macro-Average}                                                                                                                                                                               \\ \cline{3-8} 
                                                                          &                                                                     & P.                                                                   & R.                                                                   & F1                                                                  & P.                                                                   & R.                                                                   & F1                                                                  \\ \hline
BERT                                                                      & 55.50                                                               & 66.41                                                               & 45.20                                                               & 53.79                                                               & 57.47                                                               & 57.26                                                               & 55.44                                                               \\
RoBERTa                                                                   & 51.30                                                               & 60.84                                                               & 22.33                                                               & 32.66                                                               & 65.29                                                               & 22.33                                                               & 33.28                                                               \\ \hline
ResNet                                                                    & 52.30                                                               & 59.75                                                               & 51.30                                                               & 55.21                                                               & 52.42                                                               & 52.47                                                               & 52.10                                                               \\
ViT                                                                       & 53.60                                                               & 60.42                                                               & 55.14                                                               & 57.66                                                               & 53.27                                                               & 53.33                                                               & 53.17                                                               \\ \hline
Res-BERT                                                                  & 58.10                                                               & 66.17                                                               & 54.97                                                               & 60.05                                                               & 58.47                                                               & 58.63                                                               & 57.99                                                               \\
Att-BERT                                                                  & 58.30                                                               & 67.56                                                               & 52.35                                                               & 58.99                                                               & 59.23                                                               & 59.31                                                               & 58.29                                                               \\
MILNet                                                                    & 56.20                                                               & 66.83                                                               & 46.42                                                               & 54.79                                                               & 57.96                                                               & 57.79                                                               & 56.10                                                               \\
DMSD-CL                                                                   & 60.60                                                               & 64.09                                                               & 71.02                                                               & 67.38                                                               & 59.30                                                               & 58.81                                                               & 58.82                                                               \\
 \cellcolor{gray!20}\textbf{MICL} & \cellcolor{gray!20}\textbf{68.70} & \cellcolor{gray!20}\textbf{70.70} & \cellcolor{gray!20}\textbf{77.48} & \cellcolor{gray!20}\textbf{73.94} & \cellcolor{gray!20}\textbf{68.01} & \cellcolor{gray!20}\textbf{67.20} & \cellcolor{gray!20}\textbf{67.38} \\ \Xhline{1.2pt}
\end{tabular}

}
\end{table}

\subsection{Main Results}
\label{ab}
The main results are shown in Table \ref{tab:main}. %
Our analysis yields the following insights: 
\textbf{1)} The proposed MICL emerges as the most effective model, outperforming all baseline models. It records improvements ranging from 2.71\% to 5.16\% over the latest DMSD-CL model across various metrics and consistently surpasses the state-of-the-art model G$^2$SAM in all metrics. \textbf{2)}  Text-based models demonstrate superior performance over image-based models, with the RoBERTa model achieving an accuracy of 85.51\%, compared to only 67.83\% by the ViT model. This indicates that text carries a higher information density than images in the multimodal sarcasm detection task. The substantial disparity in performance causes multimodal models to rely excessively on textual data, potentially compromising their ability to generalize. These insights underscore MICL's proficiency in leveraging multimodal data to achieve exceptional results in the multimodal sarcasm detection task.

\subsection{Analysis on SPMSD}
We design a comparative experiment on the spurious correlation test set SPMSD, as shown in Table \ref{tab:ood}. The analysis reveals that, unlike the main experimental results with high recall, most baseline models exhibit lower recall compared to precision. This discrepancy in performance metrics highlights the significant impact of varying data distributions on the decision-making processes of existing models, tentatively affirming the presence of the spurious correlation issue. Notably, the proposed MICL significantly outperforms all baselines, achieving a 68.7\% accuracy rate. Specifically, against DMSD-CL, MICL displays a more significant 6.46\% to 8.71\% improvement across various metrics, which is more significant than that on MMSD. These results demonstrate that MICL can effectively mitigate reliance on spurious correlations, showing better generalization ability on new data.

\begin{table}[htbp]
\centering
\caption{Experiment results of ablation study.}
\label{tab:ab}
\resizebox{\linewidth}{!}{  
\renewcommand\arraystretch{1}
\begin{tabular}{ccccc|cccc}
\Xhline{1.2pt}
\multirow{2}{*}{\textit{Base}} & \multirow{2}{*}{$f_w$} & \multirow{2}{*}{$f_e$} & \multirow{2}{*}{$f_s$} & \multirow{2}{*}{$c$} & \multicolumn{2}{c}{MMSD}        & \multicolumn{2}{c}{SPMSD}       \\ \cline{6-9} 
                               &                        &                        &                        &                      & Acc.(\%)        & F1(\%)         & Acc.(\%)        & F1(\%)         \\ \hline
$\checkmark$                   &                        &                        &                        &                      & 88.54          & 85.73          & 61.60          & 63.77          \\
$\checkmark$                   & $\checkmark$           &                        &                        &                      & 89.97          & 87.20              & 62.80          & 65.46       \\
$\checkmark$                   & $\checkmark$           & $\checkmark$           &                        &                     & 91.32          & 89.33        & 63.70          & 65.79          \\
$\checkmark$                   & $\checkmark$           &                        & $\checkmark$           &                            & 90.77          & 88.84       & 66.10          & 69.29         \\
$\checkmark$                   & $\checkmark$           & $\checkmark$           & $\checkmark$           &                      & 91.45          & 89.51         & 67.90          & 71.23          \\ \hline
$\checkmark$                   & $\checkmark$           & $\checkmark$           & $\checkmark$           & $\checkmark$         & \textbf{92.08} & \textbf{90.33} & \textbf{68.70} & \textbf{73.94} \\ \Xhline{1.2pt}
\end{tabular}}
\end{table}

\begin{table}[htbp]

\caption{Results of using different extra data (Acc \%). $^*$ MILNet removes the OCR module, DMSD-CL removes the data augmentation module.}
\label{tab:ocr}
\resizebox{\linewidth}{!}{  
\renewcommand\arraystretch{1}
\begin{tabular}{lcc|lcc}
\Xhline{1.2pt}
Method                           & MMSD           & SPMSD          & Method                           & MMSD           & SPMSD          \\ \hline
MILNet$^*$                       & 89.54          & 56.70          & MILNet                           & 89.50          & 56.20          \\
+ \textit{OCR'} & 89.50          & 56.20          & + \textit{aug'} & 89.41          & 59.00          \\
+ \textit{ours}  & \textbf{89.66} & \textbf{57.80} & + \textit{ours}  & \textbf{89.58} & \textbf{65.40} \\ \hline
DMSD-CL                          & 88.95          & 60.60          & DMSD-CL$^*$                      & 89.08          & 57.20          \\
+ \textit{OCR'} & 88.62          & 59.10          & +  \textit{aug'} & 88.95          & 60.60          \\
+ \textit{ours}  & \textbf{89.04} & \textbf{60.90} & + \textit{ours}  & \textbf{89.29} & \textbf{65.30} \\ \hline
MICL                             & 91.40          & 67.40          & MICL                             & 91.91          & 56.90          \\
+ \textit{OCR'} & 90.27          & 64.80          & + \textit{aug'} & 91.07          & 59.20          \\
+ \textit{ours}  & \textbf{92.08} & \textbf{68.70} & + \textit{ours}  & \textbf{92.08} & \textbf{68.70} \\ \Xhline{1.2pt}
\end{tabular}

}
\end{table}

\subsection{Ablation Study}
\paragraph{Analysis of components.} To probe the effectiveness of each component in MICL, we conduct ablation experiments. The experimental results are shown in Table \ref{tab:ab}, where \textit{Base} represents the direct concatenation of $\boldsymbol{H}_v$ and $\boldsymbol{H}_t$ for prediction. $\boldsymbol{f}_w$, $\boldsymbol{f}_e$, and $\boldsymbol{f}_s$ correspond to the token-patch, entity-object, and sentiment incongruity learning modules, respectively. $c$ represents multi-view fusion using credibility. According to the results, we have the following findings: \textbf{1)} All incongruity learning modules can improve performance compared to the base model. \textbf{2)} $\boldsymbol{f}_s$ effectively improves the model's performance on the SPMSD dataset, reducing erroneous dependence on the text. \textbf{3)} $\boldsymbol{f}_e$ significantly improves performance on the MMSD dataset, proving that entity-object incongruity is crucial in the MSD task. %
\textbf{4)} $c$ can effectively integrate features from different views and improve performance.

\paragraph{Analysis of extra data.} From a data perspective, we conduct another set of ablation experiments to validate the efficacy of our OCR-text and data augmentation. The results are shown in Tables \ref{tab:ocr}, where \textit{ours} refer to the OCR-text and data augmentation proposed in this paper, \textit{OCR'} represents the OCR-text extracted by \citet{pan2020modeling}, and \textit{aug'} refers to the data augmentation of DMSD-CL. The analysis yields several key insights: \textbf{1)} Additional data does not necessarily enhance model performance. In some instances, it may even impair the model's effectiveness due to distributional differences from the original data. For example, MILNet+\textit{aug'} slightly improves performance on the SPMSD dataset but causes a decrease on the MMSD dataset. \textbf{2)} Our OCR-text can enhance the models' performance. All methods show better results on both benchmark datasets. \textbf{3)} Our novel data augmentation approach improves model robustness in spurious correlation scenarios without compromising the baseline performance. These findings collectively affirm the effectiveness of the OCR-text and data augmentation devised in our study.

\begin{figure}[t]
  \centering
  \includegraphics[width=\linewidth]{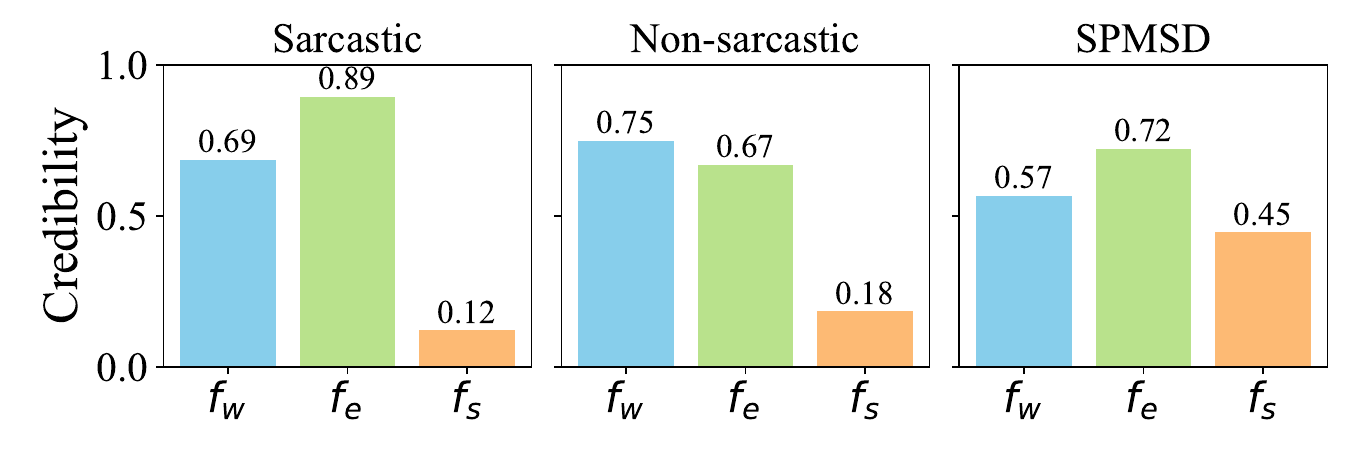}
  \caption{Credibility study.}
  \label{fig:cre}
\end{figure}

\subsection{Credibility Study}
To investigate the credibility of incongruity features from different perspectives in various scenarios, we conduct a credibility study, with the results shown in Figure \ref{fig:cre}. We divide the study into three scenarios: the sarcastic and non-sarcastic scenarios of the MMSD dataset, and the SPMSD scenario. We calculate and display the average credibility of each feature. The experimental results show that entity-object incongruity exhibits high credibility for sarcastic samples, indicating that this view is effective in capturing sarcastic entity information. Conversely, traditional token-patch incongruity effectively detects non-sarcastic samples. Moreover, sentiment incongruity is beneficial in reducing the model's dependence on spurious correlations. In addition, the credibility of each view is relatively balanced on SPMSD. Therefore, the components of our multi-view incongruity learning method complement each other across different scenarios, demonstrating effective mitigation of spurious correlation issues.

\subsection{Case Study}
\begin{figure}[t]
  \centering
  \includegraphics[width=\linewidth]{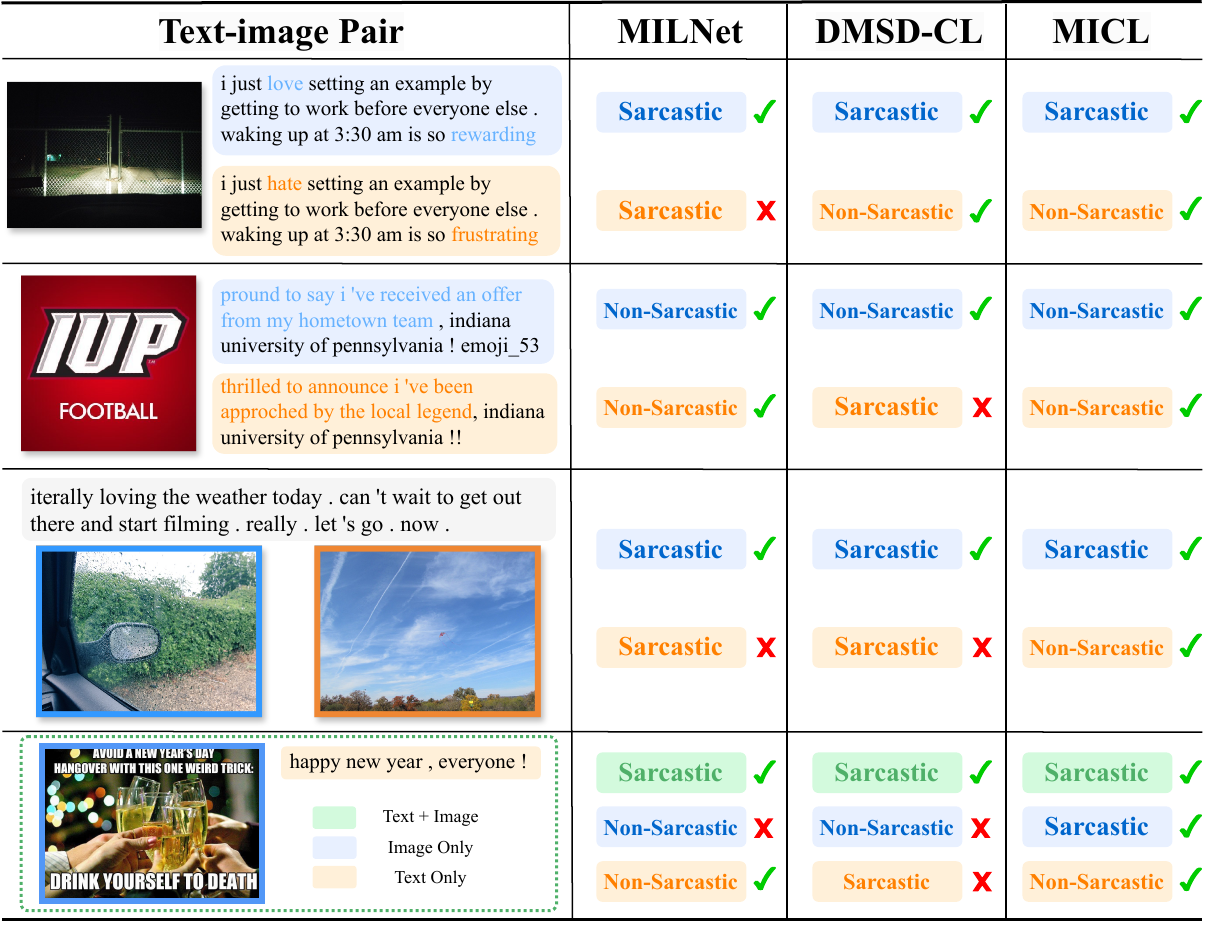}
  \caption{Case studies on spuriously correlated samples.}
  \label{fig:case}
\end{figure}

To provide an intuitive comprehension of MICL on spuriously correlated samples, we design a case study. Based on empirical summaries, we present four types of spuriously correlated samples and compare the results of  MILNet, DMSD-CL and MICL, as shown in Figure \ref{fig:case}. %
In case 1, the focus is mainly on the particular emotional words in the text. 
Case 2 investigates the impact of modifying non-critical information. Case 3 examines whether models can handle situations where the image and text are congruent. Case 4 examines whether models can correctly handle unimodal inputs. 
The results show that MILNet struggles with most spurious correlation scenarios (case 1, 3, and 4), showing obvious over-focusing on the text modality.
DMSD-CL can handle scenarios involving emotive words (case 1), but it also has modality learning bias (case 3 and 4). In addition, DMSD-CL makes mistakes in learning key textual content (case 2). Therefore, the problem of spurious correlations strongly affects the model's generalizability. %
Meanwhile, the proposed MICL, through data augmentation and multi-view incongruity learning, can detect sarcasm properly in various scenarios, emphasizing its generalizability and superiority in MSD.

\section{Conclusion}
In this paper, we introduce MICL, an innovative approach that leverages contrastive learning to learn multi-view incongruities. This method is designed to counteract the prevalent issue of spurious correlations observed in current MSD models. Furthermore, we tackle the challenge of models' excessive dependence on textual data by integrating a comprehensive text-image data augmentation scheme. To empirically highlight the problem of spurious correlations, we introduce a test set, SPMSD, built upon the foundational MMSD dataset. Experimental results show that MICL not only achieves state-of-the-art performance on the MSD task but also effectively mitigates spurious correlations.

\section{Limitation}
Although MICL can reduce the dependence on spurious correlations, it achieves only a 68\% accuracy rate on the SPMSD dataset, indicating still substantial scope for further enhancement. Our empirical experiments and existing literature \cite{wang2024generated} show that some spurious correlations can improve model performance, which is a point not discussed in this paper. Additionally, MICL's complexity, particularly with integrating hybrid attention and graph attention networks, may pose challenges in scalability and efficiency.

\section*{Acknowledgments}
This research was supported by the National Key R\&D Program of China (No. 2023YFC3303800). Hao Peng was supported by NSFC through grant 62322202. Prof. Philip S. Yu was supported in part by NSF under grants III-2106758, and POSE-2346158.

\bibliography{custom}

\newpage
\appendix

\section{Dataset}
\label{sec:spmsd}
The statistics of MMSD dataset is as shown in Table \ref{tab:mmsd}.

The SPMSD dataset is derived and expanded from the MMSD dataset, specifically designed to evaluate the models' reliance on spurious correlations. %
To ensure the fairness of the dataset, we randomly select 1,000 samples from the MMSD dataset and use these samples as the basis for constructing the SPMSD dataset. We employ various strategies to construct SPMSD, aiming to obtain a wide range of potential spurious correlations. These strategies include transforming the sentiment of the text, only describing the content of the image in the text, replacing entities in the text with entities appearing in the image, regenerating sarcastic text based on the image, swapping text-image pairs, and using only image/text.

\begin{table*}[t]

\caption{Additional experimental results with LVLMs.}
\label{tab:lvlm}
\resizebox{\linewidth}{!}{  
\renewcommand\arraystretch{1}
\begin{tabular}{lcccccc}
\Xhline{1.2pt}
\multirow{2}{*}{Method} & \multicolumn{3}{c}{MMSD}               & \multicolumn{3}{c}{SPMSD}              \\ \cline{2-7} 
                        & Acc(\%) & Binary-F1(\%) & Macro-F1(\%) & Acc(\%) & Binary-F1(\%) & Macro-F1(\%) \\ \hline
MiniCPM-V 2.0           & 55.95   & 43.59         & 53.73        & 53.30   & 46.75         & 52.58        \\
LLaVA 1.6               & 60.23   & 46.42         & 57.40        & 48.80   & 44.59         & 48.50        \\
VisualGLM               & 60.81   & 44.66         & 41.03        & 60.80   & 58.41         & 48.33        \\
Qwen-VL-Chat            & 45.08   & 27.01         & 43.27        & 44.20   & 38.83         & 45.08        \\
mPLUG-Owl 2             & 59.40   & 34.62         & 52.59        & 47.90   & 33.16         & 45.30        \\
ChatGPT 4               & 76.11   & 74.75         & 76.01        & \textbf{70.20}   & 66.37         & 64.21        \\ \hline
\textbf{MICL(ours)}                   & \textbf{92.08}   & \textbf{90.33}         & \textbf{91.81}        & 68.70   & \textbf{73.94}         & \textbf{67.38}        \\ \Xhline{1.2pt}
\end{tabular}
}
\end{table*}

\begin{table}[ht]
\centering
\caption{Statistics of MMSD.}
\label{tab:mmsd}
\begin{tabular}{lccc}
\Xhline{1.2pt}
\textbf{Label} & \textbf{Train} & \textbf{Val} & \textbf{Test} \\
\hline
Positive & 8642 & 959 & 959 \\
Negative & 11174 & 1451 & 1450 \\
All & 19816 & 2410 & 2409 \\
\Xhline{1.2pt}

\end{tabular}
\end{table}

\section{Estimating Credibility}
\label{sec:cre}
In the context of multi-class classification, Subjective logic (SL) associates the parameters of the Dirichlet distribution. Subjective logic defines a theoretical framework for obtaining the probabilities of different categories (belief masses) and the overall uncertainty (uncertainty mass) of multi-classification problems based on \textit{evidence} collected from the data. Specifically, for the $K$ classification problems, subjective logic tries to assign a belief mass to each class label and an overall uncertainty mass to the whole frame based on the \textit{evidence}. Accordingly, for the $v$-th view, the $K + 1$ mass values are all non-negative and their sum is one:
\begin{equation}
    u^v+\sum^K_{k=1}b_k^v=1,
\end{equation}
where $u^v_k>=0$ and $b_k^v>=0$ indicate the overall uncertainty and the probability for the $k$-th class, respectively.

For the $v$-th view, subjective logic connects the \textit{evidence} $\boldsymbol{e}^v=[e_1^v,...,e_K^v]$ to the parameters of the Dirichlet distribution $\boldsymbol{\alpha^v}=[\alpha_1^v,...,\alpha_K^v]$. Specifically, the parameter $\alpha_k^V$ of the Dirichlet distribution is induced from $e_k^v$, $i.e.$, $\alpha_k^v=e_k^v+1$. Then, the belief mass $b_k^v$ and the uncertainly $u^v$ are computed as:
\begin{equation}
    b_k^v = \frac{e_k^v}{S_v}=\frac{\alpha-1}{S_v}, \; u^v = \frac{K}{S_v},
\end{equation}
where $S_v = \sum_{i=1}^K(e_i^v+1)=\sum_{i=1}^K\alpha_i^v$ is the Dirichlet strength. We follow the work of \citet{ma2024event} and simply use 1 minus the uncertainty $u^v$ to estimate the credibility of each view, that is:
\begin{equation}
    \begin{split}
          c^v &= 1-u^v \\
        &=\sum^K_{k=1}b_k^v \\
        &=b_0^v+b_1^v \\
        &=\frac{e_0^v}{S_v}+\frac{e_1^v}{S_v} \\ 
        &=\frac{e_0^v+e_1^v}{(e_0^v+1)+(e_1^v+1)} .
    \end{split}
\end{equation}

\begin{table}[t]

\caption{Additional experimental results with BERT text encoder.}
\label{tab:bert}
\resizebox{\linewidth}{!}{  
\renewcommand\arraystretch{1}
\begin{tabular}{lccccccc}
\Xhline{1.2pt}
\multirow{2}{*}{Method} & \multirow{2}{*}{Acc} & \multicolumn{3}{c}{Binary-Average} & \multicolumn{3}{c}{Macro-Average} \\ \cline{3-8} 
                        &                      & P          & R         & F1        & P         & R         & F1        \\ \hline
BERT                    & 83.85                & 78.72      & 82.27     & 80.22     & 81.31     & 80.87     & 81.09     \\
Res-BERT                & 84.80                & 77.80      & 84.15     & 80.85     & 78.87     & 84.46     & 81.57     \\
Att-BERT                & 86.05                & 78.63      & 83.31     & 80.90     & 80.87     & 85.08     & 82.92     \\
MILNet                  & 88.72                & 84.97      & 87.79     & 86.37     & 87.75     & 88.29     & 88.04     \\
DMSD-CL                 & 88.24                & 86.47      & 84.42     & 85.43     & 87.65     & 87.94     & 87.79     \\
G$^2$SAM & 90.48                & 87.95      & \textbf{89.02}     & 88.48     & 89.44     & 89.79     & 89.65     \\ 

\hline
MICL(ours)              & \textbf{91.36}                & \textbf{89.48}      & 88.84     & \textbf{89.16}    & \textbf{90.90}     & \textbf{90.80}     & \textbf{90.85}     \\ 

\Xhline{1.2pt}
\end{tabular}}
\end{table}

\section{Experiments Compared with LVLMs}
Large Vision-Language Models (LVLMs) have demonstrated remarkable results across various multimodal tasks. We compare the performance of MICL with existing LVLMs on the MSD task, and the results are presented in Table \ref{tab:lvlm}. It can be seen that without fine-tuning most LVLMs
do not reach the performance of mainstream methods on the MMSD and SPMSD datasets. However, ChatGPT-4's accuracy on the SPMSD dataset is slightly higher than that of MICL.

\section{Experiments on Different Backbones}
To ensure a fair comparison of results, we standardize the text encoder of all models to BERT and conduct experiments on the MMSD dataset. The results are presented in Table \ref{tab:bert}. As shown in the table, our MICL model continues to achieve the best performance.

\section{Attention Visualization}
\begin{figure}[htbp]
  \centering
  \includegraphics[width=\linewidth]{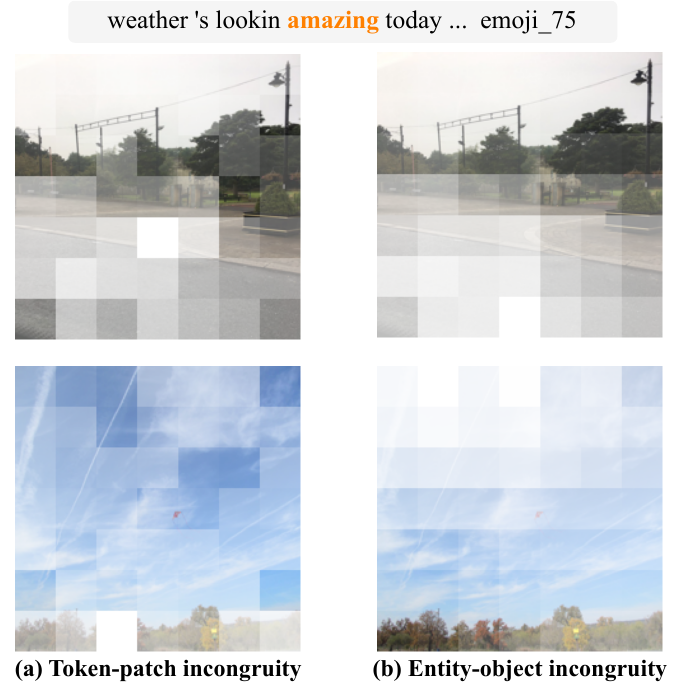}
  \caption{Attention Visualization.}
  \label{fig:att}
\end{figure}

To intuitively demonstrate the concerns of token-patch incongruity and entity-object incongruity learning, we conduct attention visualization experiments, using sub-modules with text as \textit{Query} and images as \textit{Key} and \textit{Value}. Figure \ref{fig:att} shows that in the sarcastic examples, both methods can focus on the key parts. In non-sarcastic examples, the two methods are complementary properties to learn features more comprehensively.

\section{Implementation Details}
We use the pre-trained RoBERTa-base\footnote{\url{https://huggingface.co/FacebookAI/roberta-base}} model for text encoding and the pre-trained vit-base-patch32-224\footnote{\url{https://huggingface.co/google/vit-base-patch32-224-in21k}} model for image encoding. For textual graph, we use the \verb|en_core_web_trf| model in spacy to extract dependencies between entities. For visual graphs, we add an edge between regions with cosine similarity > 0.6. We use gpt3.5-turbo for text data augmentation. For image data augmentation, we extract the original image content with GLM-4V and complete the text-to-image and image-to-image steps using stable diffusion\footnote{\url{https://huggingface.co/stabilityai/stable-diffusion-xl-base-1.0}}. We set the feature dimension $d$ to 768, and set the hyperparameters $\tau$ and $\lambda$ to 0.07 and 1, respectively. We use the Adam optimizer to optimize our model. The learning rate is set to 1e-5 for all components. The learning rate is reduced to 0 in the line schedule. All experiments are completed under a single Nvidia RTX 4090 (24 G).

\section{Prompts}
\paragraph{Prompts for OCR.}
Please perform OCR on this image and translate any non-English text into English.

\paragraph{Prompts for Text Augmentation.}
Please rewrite these data from three aspects: 1. Reverse the meaning of sarcasm: that is, if the sarcasm item of the original sarcasm data is yes, please rewrite the original text into a sentence that does not contain sarcasm at all; if the sarcasm item of the original sarcasm data is no, please use a strong sarcasm emotion rewrite text; 2. Keep the sarcasm meaning: keep the sarcasm items of the original data unchanged, introduce some new concepts, and rewrite them.

\paragraph{Prompts for Image Captioning.}
Please describe the main content of this image in one sentence.

\section{OCR-text Examples}
\begin{figure}[h]
  \centering
  \includegraphics[width=\linewidth]{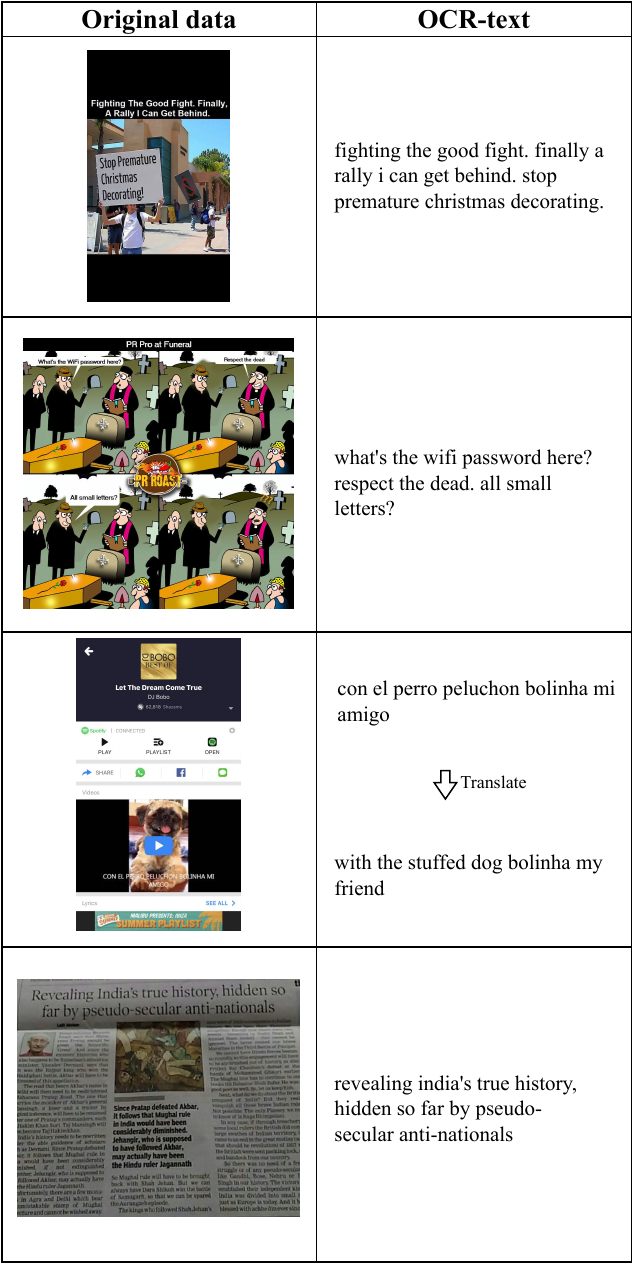}
  \caption{OCR-text examples.}
  \label{fig:app_ocr}
\end{figure}

We give more OCR-text examples, as shown in Figure \ref{fig:app_ocr}. Our approach can handle handwriting, comics, non-English, and photos.

\newpage

\section{Data Augmentation Examples}
\begin{figure}[h]
  \centering
  \includegraphics[width=\linewidth]{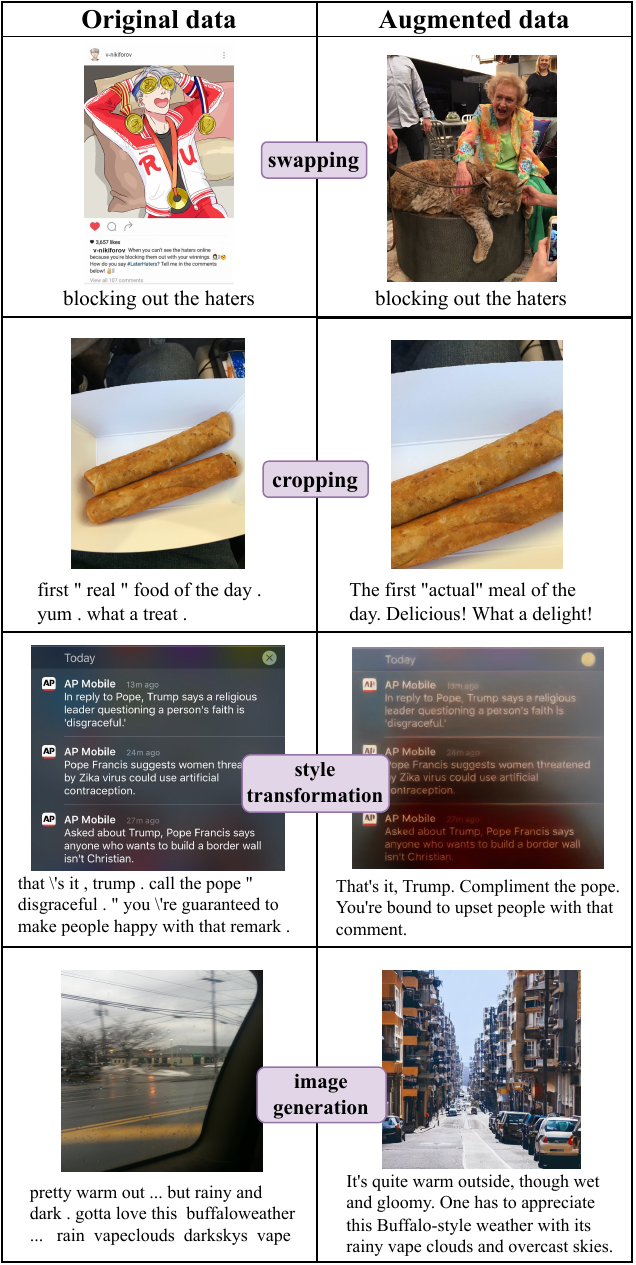}
  \caption{Data augmentation examples.}
  \label{fig:app_aug}
\end{figure}

We give more data augmentation examples, as shown in Figure \ref{fig:app_aug}.

\end{document}